%% file: paper.tex
\documentclass[conference, a4paper]{IEEEtran}
\usepackage{cite}
\usepackage[pdftex]{graphicx}
\usepackage[cmex10]{amsmath}
\usepackage{algorithm}
\usepackage{algorithmicx}
\usepackage{algpseudocode}
\usepackage{mdwmath}
\usepackage{mdwtab}
\usepackage{amssymb}
\usepackage{amsmath}
\usepackage{subfigure}
\usepackage{tabularx}
\usepackage{mdwlist}
\usepackage{url}
\usepackage{multirow}

\ifCLASSINFOpdf
\else
\fi

\newcommand{\comment}[1]{}

\DeclareMathAlphabet{\mathitbf}{OML}{cmm}{b}{it}

\newcommand{\eg}{\emph{e.g.}}
\newcommand{\ie}{\emph{i.e.}}
\newcommand{\etal}{\emph{et~al.}}
\renewcommand{\Huge}{\fontsize{23.5pt}{27pt}\selectfont}
\newcommand{\sq}{\hbox{\rlap{$\sqcap$}$\sqcup$}}
\newcommand{\qed}{\hspace*{\fill}\sq}

\begin{document}
\title{Evaluation of BERT and ALBERT Sentence Embedding Performance on Downstream NLP Tasks}

\author{
\IEEEauthorblockN{Hyunjin Choi, Judong Kim, Seongho Joe, and Youngjune Gwon}
\IEEEauthorblockA{Samsung SDS, Seoul, Korea}
}

\maketitle

\begin{abstract}
Contextualized representations from a pre-trained language model are central to achieve a high performance on downstream NLP task. The pre-trained BERT and A Lite BERT (ALBERT) models can be fine-tuned to give state-of-the-art results in sentence-pair regressions such as semantic textual similarity (STS) and natural language inference (NLI). Although BERT-based models yield the \texttt{[CLS]} token vector as a reasonable sentence embedding, the search for an optimal sentence embedding scheme remains an active research area in computational linguistics. This paper explores on sentence embedding models for BERT and ALBERT. In particular, we take a modified BERT network with siamese and triplet network structures called Sentence-BERT (SBERT) and replace BERT with ALBERT to create Sentence-ALBERT (SALBERT). We also experiment with an outer CNN sentence-embedding network for SBERT and SALBERT. We evaluate performances of all sentence-embedding models considered using the STS and NLI datasets. The empirical results indicate that our CNN architecture improves ALBERT models substantially more than BERT models for STS benchmark. Despite significantly fewer model parameters, ALBERT sentence embedding is highly competitive to BERT in downstream NLP evaluations. 
\end{abstract}

\IEEEpeerreviewmaketitle

% sections
\input{intro}
\input{related}
\input{models}

\input{exp}
\input{conc}

% references
{
\bibliographystyle{IEEEtran}
\bibliography{paper}
}
\end{document}

%% file: intro.tex
\section{Introduction}
Pre-trained language models have impacted the way modern natural language processing (NLP) applications and systems are built. An important paradigm is to train a language model on large corpora to serve as a platform upon which an NLP application can be built and optimized. Such platform is shareable and can be distributed. Self-supervised learning with large corpora provides an appropriate starting point for extra task-specific layers being optimized from scratch while reusing the pre-trained model parameters. 

Transformer~\cite{transformer}, a sequence transduction model based on attention mechanism, has revolutionized the design of a neural encoder for natural language sequences. By skipping any recurrent or convolutional structures, the transformer
architecture enables the learning of sequential information in an input solely via attention, thanks to multihead self-attention layers in an encoder block. Devlin~\etal~\cite{bert} have proposed Bidirectional Encoder Representations from Transformers (BERT) to improve on predominantly unidirectional training of language models. 

By jointly conditioning on both left and right context in all layers, BERT uses the masked language modeling (MLM) loss to make the training of deep bidirectional
language encoding possible. BERT uses an additional loss for pre-training known as next-sentence prediction (NSP). NSP is designed to learn high-level linguistic coherence by predicting whether or not given two text segments should appear consecutively as in the original text. NSP is expected to improve downstream NLP task performances such as semantic textual similarity (STS) and natural language inference (NLI) that need to infer reasoning about inter-sentence relations.

A Lite BERT~\cite{albert} is proposed to scale up the language representation learning via parameter reduction techniques. In ALBERT, cross-layer parameter sharing and factorization of embedding parameters can be thought as a regularization that helps stabilize its training. Furthermore, ALBERT uses an updated self-supervised loss known as sentence-order prediction (SOP) that enhances the ineffectiveness of NSP confused between topic and coherence predictions. SOP has been shown to consistently help downstream tasks with multi-sentence inputs.

The pre-training tasks are intrinsic compared to downstream tasks. A key disadvantage of BERT is that no independent sentence embeddings are computed. As a higher means of abstraction, sentence embeddings can play a central role to achieve good downstream performances like machine reading comprehension (MRC). 

The specifics of NLP applications are well-abstracted by downstream tasks. For this reason, downstream performance is a good indicator for a language model. When pre-trained language models are used for downstream task evaluations, pre-trained models can generate additional feature representations in addition to being provided as a platform for fine-tuning. 

In this paper, we are interested in learning sentence representation using out-of-the-box BERT and ALBERT token embeddings. Sentence embedding models are essential for clustering and semantic search where a sentence input is mapped in a high-dimensional semantic vector space such that sentence vectors with similar meanings are close in distance. NLP researchers have started to input an individual sentence into BERT to derive a fixed-size embedding. A commonly accepted sentence embedding for BERT-based models is the \texttt{[CLS]} token used for sentence-order prediction (\ie, NSP or SOP) during the pre-training. 

Averaging the representations obtained from the BERT or ALBERT output layer (\ie, token embeddings) gives an alternative. Using the \texttt{[CLS]} token, which is optimized by an intrinsic task of the pre-training, is considered suboptimal while the average pooling of token embeddings has a limitation of its own. Nonetheless, it can be time consuming to perform multi-sentence tasks associated with semantic search, summarization, and paraphrase. 

Computing sentence embeddings from contextualized language models is an active, ongoing research problem. In our exploration for more elaborate sentence embedding models, we first consider Sentence-BERT (SBERT)~\cite{sentencebert}, a modified BERT network with siamese and triplet network structures to derive semantically meaningful sentence embeddings. SBERT is computationally efficient and can compare sentences using only cosine-similarity at run-time. We then take the SBERT architecture and simply replace BERT with ALBERT to form Sentence-ALBERT (SALBERT). We also apply a convolutional neural net (CNN) instead of average pooling that takes in the BERT or ALBERT token embedding outputs. 

We have evaluated the empirical performance of all sentence embedding models by using the STS and NLI datasets. We find that our CNN architecture improves ALBERT models up to 8 points in Spearman's rank correlation for STS benchmark, which is substantially larger than the case for BERT models with an improvement of only 1 point. Despite significantly fewer model parameters, ALBERT sentence embedding is highly competitive to BERT in downstream NLP evaluations.

This paper is structured in the following manner. Section II presents related work. Section III describes all sentence embedding models of our consideration. In Section IV, we empirically evaluate the sentence embedding models using the STS and NLI datasets. The paper concludes in Section V. 

%% file: related.tex
\section{Related Work}
Language models provide core building blocks for downstream NLP tasks. Task-specific fine-tuning of a pre-trained language model is a contemporary approach to implement an NLP system. BERT~\cite{bert} is a pre-trained transformer encoder network~\cite{transformer} fine-tuned to give state-of-the-art results in question answering, sentence classification, and sentence-pair regression. A Lite BERT (ALBERT)~\cite{albert} incorporates parameter reduction techniques to scale better than BERT. ALBERT is known to improve on inter-sentence coherence by a self-supervised loss from sentence-order prediction (SOP) compared to the next sentence prediction (NSP) loss in the original BERT. 

The BERT network structure contains a special classification token \texttt{[CLS]} as an aggregate sequence representation for NSP. (Similarly for ALBERT, \texttt{[CLS]} is used for SOP.) The \texttt{[CLS]} token therefore can serve a sentence embedding. Because there are no other independently computed sentence embeddings for BERT and ALBERT, one can average-pool the token embedding outputs to form a fixed-length sentence vector. 

Previously, sentence embedding research looked over convolutional and recurrent structures as building blocks. Kim~\cite{cnnforsc} proposed a CNN with max pooling for sentence classification. In Conneau \etal~\cite{conneau2017supervised}, bidirectional LSTM (BiLSTM) was used as sentence embedding for natural language inference tasks. More complex neural nets such as Socher~\etal~\cite{socher-etal-2013-recursive} introduced recursive neural tensor network (RNTN) over parse trees to compute sentence embedding for sentiment analysis. Zhu~\etal~\cite{zhu2015long} and Tai~\etal~\cite{tai2015improved} proposed tree-LSTM while Yu \& Munkhdalai~\cite{munkhdalai2016neural} suggested neural semantic encoder (NSE) based on memory augmented neural net. 

Recently, sentence embedding research is exploring attention mechanisms. Vaswani~\etal~\cite{transformer} have proposed Transformer, a self-attention network for the neural sequence-to-sequence task. A self-attention network uses multi-head scaled dot-product attention to represent each word as a weighted sum of all words in the sentence. The idea of self-attention pooling has existed before self-attention network as in Liu~\etal~\cite{liu2016learning} that have utilized inner-attention within a sentence to apply pooling for sentence embedding. Choi~\etal~\cite{choi2018finegrained} have developed a fine-grained attention mechanism for neural machine translation, extending scalar attention to vectors. 

Complex contextualized sentence encoders are usually pre-trained like language models, but they can be improved by supervised transfer tasks such as natural language inference (NLI). InferSent by Conneau~\etal~\cite{conneau2017supervised} has consistently outperformed unsupervised methods like SkipThought. Universal Sentence
Encoder~\cite{use} trains a transformer network and augments unsupervised learning with training on the Stanford NLI (SNLI) dataset. Hill~\etal~\cite{hill-etal-2016-learning} show that the task on which sentence embeddings are trained significantly impacts their quality. According to Conneau~\etal~\cite{conneau2017supervised} and Cer~\etal~\cite{use}, the SNLI datasets are suitable for training sentence embeddings. Yang~\etal~\cite{yang-etal-2018-learning} present a method to train siamese deep averaging network (DAN) and transformer, using conversations from Reddit to yield good results on the STS benchmark. 

In Sentence-BERT (SBERT)~\cite{sentencebert}, a comprehensive evaluation on the pre-trained BERT combined with siamese and triplet network structures is presented. To alleviate the run-time overhead, SBERT's more elaborate fine-tuning mechanisms such as softmax on augmented sentence representations and triplet loss are replaced by the cosine similarity at inference. The simplistic SBERT inference helps reduce the effort for finding the most similar pair from 65 hours with BERT to about 5 seconds, while hardly impacting the accuracy. 

%% file: models.tex
\section{Models}
The output of BERT or ALBERT constitutes token embeddings for a given text input. With a large output size (\eg, up to 512 token vectors of 768 dimensions each), the contextualized word embeddings can be fine-tuned for any downstream task. To do sentence-level regressions such as semantic textual similarity (STS), fixed-size sentence embeddings would be necessary. In this section, we describe sentence embedding models for BERT and ALBERT. 

%[Goldberg (2019); Hewitt and Manning (2019); Liu et al. (2019); Tenney et al. (2019)] Jawahar, Sagot \& Seddah (2019) show that the BERT layers learn different levels of information and linguistic properties. Phrase-level representations are believed to be conveyed in the lower layers. The intermediate layers encode a hierarchy of linguistically rich features that are transferable, and the most semantically expressive features at the top.

\subsection{The \texttt{[CLS]} token embedding}
The most straightforward sentence embedding model is the \texttt{[CLS]} vector used to predict sentence-level context (\ie, BERT NSP, ALBERT SOP) during the pre-training. The \texttt{[CLS]} token summarizes the information from other tokens via a self-attention mechanism that facilitates the intrinsic tasks of the pre-training. A similar reasoning applies such that the \texttt{[CLS]} token can be further optimized while fine-tuning the downstream task. After the fine-tuning, the \texttt{[CLS]} token is expected to capture more semantically-relevant sentence-level context specific to the downstream task.

\subsection{Pooled token embeddings}
Averaging the token embedding output gives our next model. The model works like a pooling layer in a convolutional neural net. Average pooling turns the token embeddings into a fixed-length sentence vector. An alternative would use max pooling instead, although max pooling tends to select the most important features rather than taking representative summary. In this paper, we choose to go with the average-pooling model. 

\subsection{Sentence-BERT (SBERT)}
Reimers \& Gurevych~\cite{sentencebert} propose SBERT that modifies a pre-trained BERT with siamese and triplet network structures to derive semantically meaningful sentence embeddings comparable using only cosine similarity. The siamese architecture is computationally efficient. Note that using a single copy of pre-trained BERT would require to run all possible combinations of sentence pairs from a dataset to form a representation for sentence pairs. SBERT first average-pools a pair of the BERT embeddings to fixed-size sentence embeddings. Using the two sentence embeddings and an element-wise difference between them, SBERT can run a softmax layer configured for classification and regression tasks.

\subsection{Sentence-ALBERT (SALBERT)}
\begin{figure}[h]
\centering
\includegraphics[width=.33\textwidth]{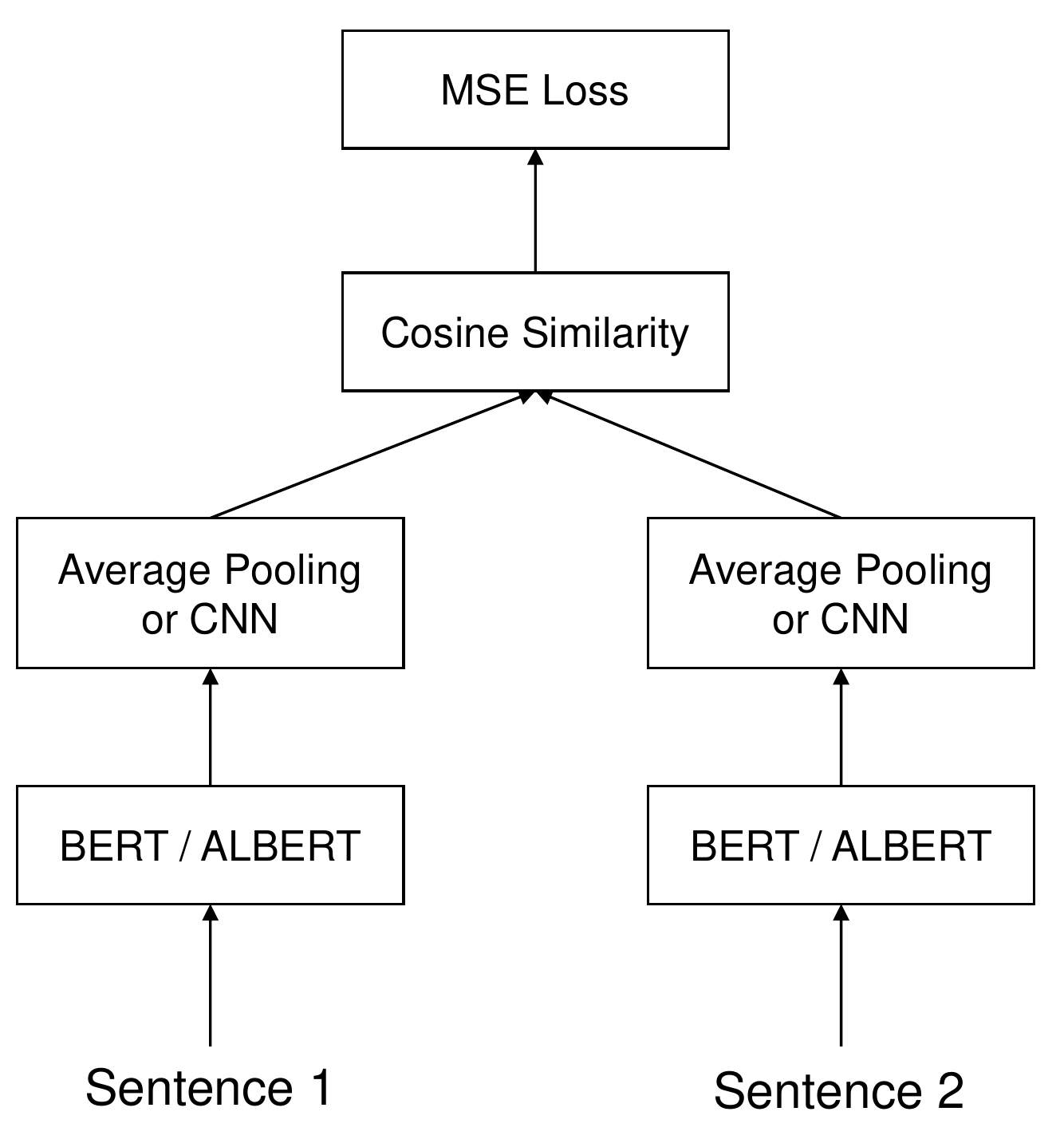}
\caption{Siamese network structure used in SBERT and SALBERT}
\label{fig:siam}
\end{figure}

Based on ALBERT, SALBERT has the same siamese and triplet networks as SBERT. The siamese network structure in SBERT and SALBERT is illustrated in Fig.~\ref{fig:siam}.

%\begin{figure}[h]
%\centering
%\includegraphics[width=.33\textwidth]{fig/siamese2}
%\caption{Siamese network structure used in SBERT and SALBERT for the NLI tasks}
%\label{fig:siam2}
%\end{figure}

\subsection{CNN-SBERT}
\begin{figure}[h]
\centering
\includegraphics[width=.2\textwidth]{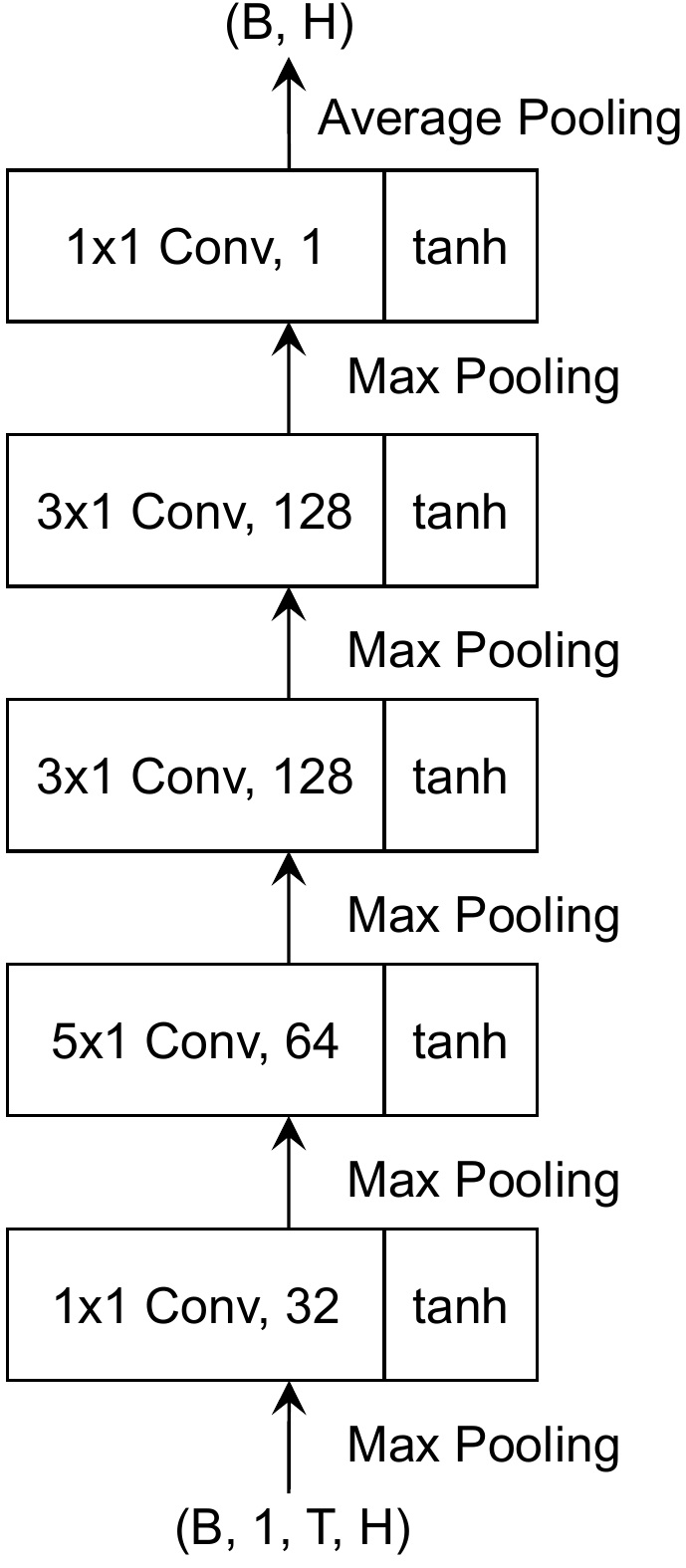}
\caption{CNN architecture used in CNN-SBERT and CNN-SALBERT. B, T, and H means mini-batch size, number of tokens, and transformer hidden size.}
\label{fig:cnn}
\end{figure}

In SBERT, average pooling is used to make the BERT embeddings into fixed-length sentence vectors. CNN-SBERT instead employs a CNN architecture that takes in the token embeddings and computes a fixed-size sentence embedding through convolutional layers with the hyperbolic tangent activation function interlaced with pooling layers. In CNN-SBERT, all the pooling layers use max pooling except the final average pooling. The CNN architecture used in CNN-SBERT is described in Fig.~\ref{fig:cnn}. 

\subsection{CNN-SALBERT}
Similarly, CNN-SALBERT uses the same CNN architecture used in CNN-SBERT.

%% file: exp.tex
\section{Experiments}
We evaluate the performance of the sentence embedding models on Semantic Textual Similarity (STS) and Natural Language Inference (NLI) benchmarks. Following the methodology by Reimers \& Gurevych~\cite{sentencebert}, we use cosine-similarity as a main metric to evaluate the similarity between two sentence embeddings. We compute both Pearson and Spearman's rank coefficients to indicate how our cosine-similarity estimate and a ground-truth label provided by the datasets are correlated. We use pre-trained BERT and ALBERT models from Hugging Face~\cite{huggingface}\footnote{\url{https://github.com/huggingface}}. 

\subsection{Datasets and tasks}
We fine-tune the BERT and ALBERT sentence embedding models on the Semantic Textual Similarity benchmark (STSb)~\cite{stsb}, the Multi-Genre Natural Language Inference (MultiNLI)~\cite{mnli}, and the Stanford Natural Language Inference (SNLI)~\cite{snli} datasets. 

\subsubsection{Semantic Textual Similarity benchmark}
STSb gives a set of English data used for STS tasks organized in International Workshop on Semantic Evaluation (SemEval)~\cite{semeval} between 2012 and 2017. The dataset includes 8,628 sentence pairs from image captions, news headlines, and user forums that are partitioned in train (5,749), dev (1,500) and test (1,379) sets. They are annotated with a score from 0 to 5 indicating how similar a pair of sentences are in terms of semantic relatedness. 

\subsubsection{Multi-genre Natural Language Inference}
The MultiNLI corpus~\cite{mnli} is a crowd-sourced collection of 433k sentence pairs annotated with textual entailment information. The dataset is used to evaluate entailment classification task. MultiNLI is modeled on the SNLI corpus, differing in its coverage of genres of spoken and written text. MultiNLI supports a distinctive cross-genre generalization evaluation. Each sentence pair in MultiNLI has a label that distinguishes whether the two sentences are contradiction, entailment, or neutral.

\subsubsection{Stanford Natural Language Inference}
The SNLI corpus~\cite{snli} contains 570k human-written English sentence pairs manually labeled for balanced classification with the labels entailment, contradiction, and neutral for natural language inference (NLI), also known as recognizing textual entailment (RTE). The General Language Understanding Evaluation (GLUE) benchmark~\cite{glue} recommends the SNLI dataset used as an auxiliary training data for MultiNLI task. Conneau~\etal~\cite{conneau2017supervised} and Cer~\etal~\cite{use} find SNLI suitable for training sentence embeddings for asserting reasoning about the semantic relationship within sentences.

\subsection{Training}
In our evaluation, we consider only BERT and ALBERT base models (\ie, multi-head attention over 12 layers) in the transformer package downloaded from Hugging Face~\cite{huggingface}. We use GLUE benchmark to fine-tune the \texttt{[CLS]} token embedding and average-pooled token embedding models with a learning rate of $3\times 10^{-5}$. We train all of our models using the Adam optimizer with a linear learning rate warm-up for 10\% of the training data. We use a learning rate of $2\times 10^{-5}$ for SBERT and SALBERT as suggested by the original SBERT architecture and $1\times 10^{-5}$ for CNN-SBERT and CNN-SALBERT. Using the MultiNLI and SNLI data, we optimize SBERT and SALBERT on the 3-way softmax loss. 

\subsubsection{STSb}
To train STS benchmark task, we use siamese network as shown in Fig~\ref{fig:siam}. We run 10 training epochs with a batch size of 32. 
\subsubsection{NLI (MultiNLI + SNLI)}
To train NLI tasks, we adopt the siamese architecture in Fig~\ref{fig:siam}. We use a softmax classifier instead of cosine similarity in training NLI tasks with a cross-entropy loss. We train 1 epoch because the NLI train set is much bigger than STSb. We use a batch size of 16.
\subsubsection{NLI + STSb}
After fine-tuning on the NLI dataset, we train on the STS benchmark with a batch size of 32.

\begin{table}[]
\centering
\caption{Evaluation on the STSb by fine-tuning sentence embeddings on STS, NLI, and both}
\label{tab:eval1}
\begin{tabular}{|l|c|l}
\cline{1-2}
\multicolumn{1}{|c|}{Model}        & Spearman (Pearson) &  \\ \cline{1-2}
\multicolumn{2}{|c|}{Not fine-tuned}                    &  \\ \cline{1-2}
BERT {[}CLS{]}-token embedding     & 6.43 (1.70)        &  \\
BERT Avg. pooled token embedding   & 47.29 (47.91)      &  \\ \cline{1-2}
ALBERT {[}CLS{]}-token embedding   & 0.86 (4.57)        &  \\
ALBERT Avg. pooled token embedding & 47.84 (46.57)      &  \\ \cline{1-2}
\multicolumn{2}{|c|}{Fine-tuned on STSb}                &  \\ \cline{1-2}
BERT {[}CLS{]}-token embedding     & 12.96 (7.49)       &  \\
BERT Avg. pooled token embedding   & 55.76 (54.90)      &  \\
SBERT                              & 84.66 (84.86)      &  \\
CNN-SBERT                          & \textbf{85.72 (86.15)}      &  \\ \cline{1-2}
ALBERT {[}CLS{]}-token embedding   & 37.98 (27.89)      &  \\
ALBERT Avg. pooled token embedding & 61.06 (60.41)      &  \\
SALBERT                            & 74.33 (75.26)      &  \\
CNN-SALBERT                        & \textbf{82.30 (83.08)}      &  \\ \cline{1-2}
\multicolumn{2}{|c|}{Fine-tuned on NLI (MultiNLI + SNLI)}                 &  \\ \cline{1-2}
BERT {[}CLS{]}-token embedding     & 32.72 (26.88)      &  \\
BERT Avg. pooled token embedding   & 69.57 (68.49)      &  \\
SBERT                              & \textbf{77.22 (74.53)}      &  \\
CNN-SBERT                          & 76.77 (75.31)      &  \\ \cline{1-2}
ALBERT {[}CLS{]}-token embedding   & 24.87 (4.11)       &  \\
ALBERT Avg. pooled token embedding & 54.21 (53.58)      &  \\
SALBERT                            & \textbf{74.05 (70.78)}      &  \\
CNN-SALBERT                        & 73.70 (72.24)      &  \\ \cline{1-2}
\multicolumn{2}{|c|}{Fine-tuned on NLI (MultiNLI + SNLI) and STSb}        &  \\ \cline{1-2}
BERT {[}CLS{]}-token embedding     & 44.77 (38.74)      &  \\
BERT Avg. pooled token embedding   & 67.61 (65.30)      &  \\
SBERT                              & 85.32 (84.51)      &  \\
CNN-SBERT                          & \textbf{85.91 (85.63)}      &  \\ \cline{1-2}
ALBERT {[}CLS{]}-token embedding   & 40.35 (33.46)      &  \\
ALBERT Avg. pooled token embedding & 60.24 (59.98)      &  \\
SALBERT                            & 77.59 (77.82)      &  \\
CNN-SALBERT                        & \textbf{83.49 (83.87)}      &  \\ \cline{1-2}
\end{tabular}
\end{table}

\begin{table}[]
\centering
\caption{Evaluation on the GLUE STSb task.}
\label{tab:eval3}
\begin{tabular}{|c|c|}
\hline
Model  & Spearman (Pearson) \\ \hline
BERT   & 88.58 (88.89)      \\ \hline
ALBERT & 90.13 (90.46)      \\ \hline
\end{tabular}
\end{table}

\begin{table*}[]
\centering
\caption{Evaluation on various STS tasks. Numbers represent Spearman (Pearson).}
\label{tab:eval2}
\begin{tabular}{|l|c|c|c|c|c|c|c|}
\hline
\multicolumn{1}{|c|}{Model} & STS12                 & STS13                 & STS14                 & STS15                 & STS16                 & STSb                  & Avg.                  \\ \hline
SBERT                       & \textbf{72.37 (77.61)} & 87.49 (88.01)          & \textbf{89.57 (89.87)} & \textbf{89.76 (89.54)} & 82.41 (80.59)          & 85.32 (84.51)          & \textbf{84.49 (85.02)} \\
CNN-SBERT                   & 69.80 (75.04)          & \textbf{88.92 (89.86)} & 89.23 (90.53)          & 89.35 (89.37)          & \textbf{82.81 (82.03)} & \textbf{85.91 (85.63)} & 84.34 (85.41)          \\ \hline
SALBERT                     & 63.87 (68.15)          & 84.04 (84.59)          & 84.89 (86.09)          & 86.41 (86.31)          & 75.26 (74.32)          & 77.59 (77.82)          & 78.68 (79.55)          \\
CNN-SALBERT                 & \textbf{65.41 (68.58)} & \textbf{86.76 (87.29)} & \textbf{86.17 (87.66)} & \textbf{87.84 (88.12)} & \textbf{81.58 (81.48)} & \textbf{83.49 (83.87)} & \textbf{81.76 (82.70)} \\ \hline
\end{tabular}
\end{table*}

\subsection{Results}
\subsubsection{Effect of fine-tuning}
Table~\ref{tab:eval1} presents the STS benchmark results. Note that performance we report is $\rho \times 100$, where $\rho$ is Spearman's rank or Pearson correlation coefficient. In general, fine-tuning results in a better performance than no fine-tuning. Without fine-tuning, the \texttt{[CLS]} token as a sentence embedding gives poor downstream task performance. Quality of sentence embedding reflected on the STSb performance seems to be affected by how related train sets used for fine-tuning are to the task. We consider STSb train set, which is directly related to the task of STSb. We also consider NLI (\ie, MultiNLI and SNLI) train sets that are not directly related to STSb. We have experimented with the following: i) fine-tuning with only STSb train set, ii) with only NLI train sets, and iii) with both NLI and STSb train sets. Fine-tuning with only STSb train set gives a reasonably good performance whereas fine-tuning with irrelevant NLI train sets only have yielded a suboptimal performance as expected. Our best STSb results are obtained by fine-tuning with both STSb and NLI train sets.

\subsubsection{Model comparison}
We expect a more elaborate sentence embedding model to give a better performance in STSb. We have found that pooling token embeddings will form a better sentence representation than \texttt{[CLS]}. We have also found that siamese structure further helps sentence embeddings. Generally, our CNN-based sentence embedding models give the best performance among all sentence embedding models.

\subsubsection{Performance of ALBERT}
ALBERT-based sentence embedding models generally achieve lower performance than the BERT counterparts in STSb evaluations. Before fine-tuning, there is no significant difference between ALBERT and BERT. The gap, however, increases after fine-tuning. Only for \texttt{[CLS]} token embedding and average-pooled embeddings, ALBERT has a better performance than BERT when fine-tuned on STSb. SALBERT has much lower performance than SBERT even though they both have the same siamese architecture. This is surprising because ALBERT has a higher score than BERT when evaluated on STSb using GLUE as shown in Table~\ref{tab:eval3}. The performance of SALBERT catches up with SBERT when the CNN architecture applies, but CNN-SALBERT is still slightly inferior to CNN-SBERT. 

\subsubsection{Effect of CNN}
In Table~\ref{tab:eval1}, we find that the best score is from CNN-based models trained on NLI and STSb. According to these scores, the CNN architecture seems to have a positive impact on sentence embedding performances. The CNN architecture, however, improves the ALBERT-based sentence embedding models more than the BERT-based models. We have found that the improvement by CNN to ALBERT models can be as high as 8 points, which is compared to 1 point for the case of BERT models. We have empirically observed that ALBERT exposes more instability (due to parameter sharing) compared to BERT. Such instability can be alleviated by CNN, and this is a possible explanation for more improvement on ALBERT by adding CNN than BERT.

\subsubsection{Evaluation of STS12--STS16 Tasks}
In Table~\ref{tab:eval2}, we present a comprehensive evaluation on the STS tasks from 2012 to 2016~\cite{2012,2013,2014,2015,2016} after fine-tuning with both the NLI and STSb train sets. We show the results of our best two models (\ie,~SBERT/SALBERT and CNN-SBERT/CNN-SALBERT). The STSb result is also presented for comparison. The purpose of the evaluation is to verify the improvement by CNN beyond STSb. In general, we see a similar trend that CNN architecture improves ALBERT-based sentence embedding models substantially more than BERT-based. On the average, SBERT embeddings achieve a Spearman's rank correlation point of 84.49 while the average for CNN-SBERT is 84.34. The CNN architecture seems almost no effect on BERT-based sentence embedding models. On the other hand, the average correlation score of CNN-SALBERT is improved by 2 points.

%% file: conc.tex
\section{Conclusion and Future Work}
In this paper, we have presented an evaluation of BERT and ALBERT sentence embedding models on Semantic Textual Similarity (STS). Knowing limitations of the \texttt{[CLS]} sentence vector, we facilitate the STS sentence-pair regression task with the siamese and triplet network architecture by Reimers \& Gurevych for BERT and ALBERT. We have additionally developed a CNN architecture that takes in the token embeddings to compute a fixed-size sentence vector. Our CNN architecture improves ALBERT models up to 0.08 (8 points in percentile) in Spearman's rank correlation for STS benchmark, which is substantially larger than the case for BERT models with an improvement of only 0.01 (1 point). Despite significantly fewer model parameters, ALBERT sentence embedding is highly competitive to BERT in downstream NLP evaluations. 

For our future work, we plan to evaluate sentence embedding with larger ALBERT models---\ie, ALBERT-large and ALBERT-xlarge. (Note that the total number of parameters in ALBERT-xlarge is still fewer than that of BERT-base.) The ALBERT results in this paper are obtained with the number of groups for the hidden layers (\texttt{num\_hidden\_groups}) set to 1. We also plan to optimize the \texttt{num\_hidden\_groups} hyperparameter for better performance.   